\documentclass[journal]{IEEEtran}
\usepackage{enumitem}
\usepackage{graphicx}
\usepackage{tabularx}
\usepackage{xcolor}
\usepackage{amsmath}
\usepackage{booktabs}
\usepackage{amssymb}
\usepackage{siunitx}
\usepackage{makecell}
\usepackage{caption}
\IEEEoverridecommandlockouts                             

\title{\LARGE \bf

Neuromorphic BrailleNet: Accurate and Generalizable Braille Reading Beyond Single Characters through Event-Based Optical Tactile Sensing}

\author{%
Naqash Afzal$^{*1}$,
Niklas Funk$^{*2}$,
Erik Helmut$^{*2}$,
Jan Peters$^{2,3}$,
and Benjamin Ward-Cherrier$^{1}$%
\thanks{$^{*}$ N. Afzal, N. Funk and E. Helmut contributed equally to this work.}%
\thanks{This work was supported by the 
Royal Academy of Engineering Fellowship on “Shared Autonomy Neuroprosthetics: Bridging Artificial and Biological Touch” under Grant RF\textbackslash 202021\textbackslash 20\textbackslash 171, the German Federal Ministry of Research, Technology and Space (BMFTR) under the Robotics Institute Germany (RIG) and the EU’s Horizon Europe project ARISE (Grant no.: 101135959).}%
\thanks{$^{1}$ N. Afzal and B. Ward-Cherrier are with the School of Engineering Mathematics and Technology, University of Bristol, BS8 1QU Bristol, U.K.
{\tt\small b.ward-cherrier@bristol.ac.uk}}%
\thanks{$^{2}$ N. Funk, E. Helmut and J. Peters are with the Department of Computer Science, Technical University of Darmstadt, Darmstadt, Germany.}%
\thanks{$^{3}$ J. Peters is also with the German Research Center for AI (DFKI), the Hessian Center for Artificial Intelligence (hessian.AI), and the Centre for Cognitive Science, Technical University of Darmstadt, Darmstadt, Germany.}%
}

\begin{document}

\maketitle
\thispagestyle{empty}
\pagestyle{empty}

\begin{abstract}

Conventional robotic Braille readers typically rely on discrete, character-by-character scanning, limiting reading speed and disrupting natural flow. Vision-based alternatives often require substantial computation, introduce latency, and degrade in real-world conditions. In this work, we present a high-accuracy, real-time pipeline for continuous Braille recognition using Evetac, an open-source neuromorphic event-based tactile sensor. Unlike frame-based vision systems, the neuromorphic tactile modality directly encodes dynamic contact events during continuous sliding, closely emulating human finger-scanning strategies.
Our approach combines spatiotemporal segmentation with a lightweight ResNet-based classifier to process sparse event streams, enabling robust character recognition across varying indentation depths and scanning speeds. The proposed system achieves near-perfect accuracy ($\geq$98\%) at standard depths, generalizes across multiple Braille board layouts, and maintains strong performance under fast scanning. On a physical Braille board containing daily-living vocabulary, the system attains over 90\% word-level accuracy, demonstrating robustness to temporal compression effects that challenge conventional methods. These results position neuromorphic tactile sensing as a scalable, low-latency solution for robotic Braille reading, with broader implications for tactile perception in assistive and robotic applications.

\end{abstract}


\section{INTRODUCTION}

Touch is one of the richest human sensory modalities, supporting fine spatial discrimination, texture perception, and dexterous object manipulation~\cite{loomis1981braille}. For individuals with visual impairments, it serves not merely as a complement to vision but as a primary channel for perceiving and interacting with the world, and is essential for both independence and agency. Replicating this capability in artificial systems remains a significant challenge. It requires tactile sensors with high spatial and temporal resolution, sensitivity to dynamic interactions, and the ability to process data efficiently and in real time. Meeting these requirements is crucial for advancing robotic dexterity as well as for developing assistive technologies that can restore or augment human touch.

\begin{figure}[htbp]
\includegraphics[width=\linewidth]{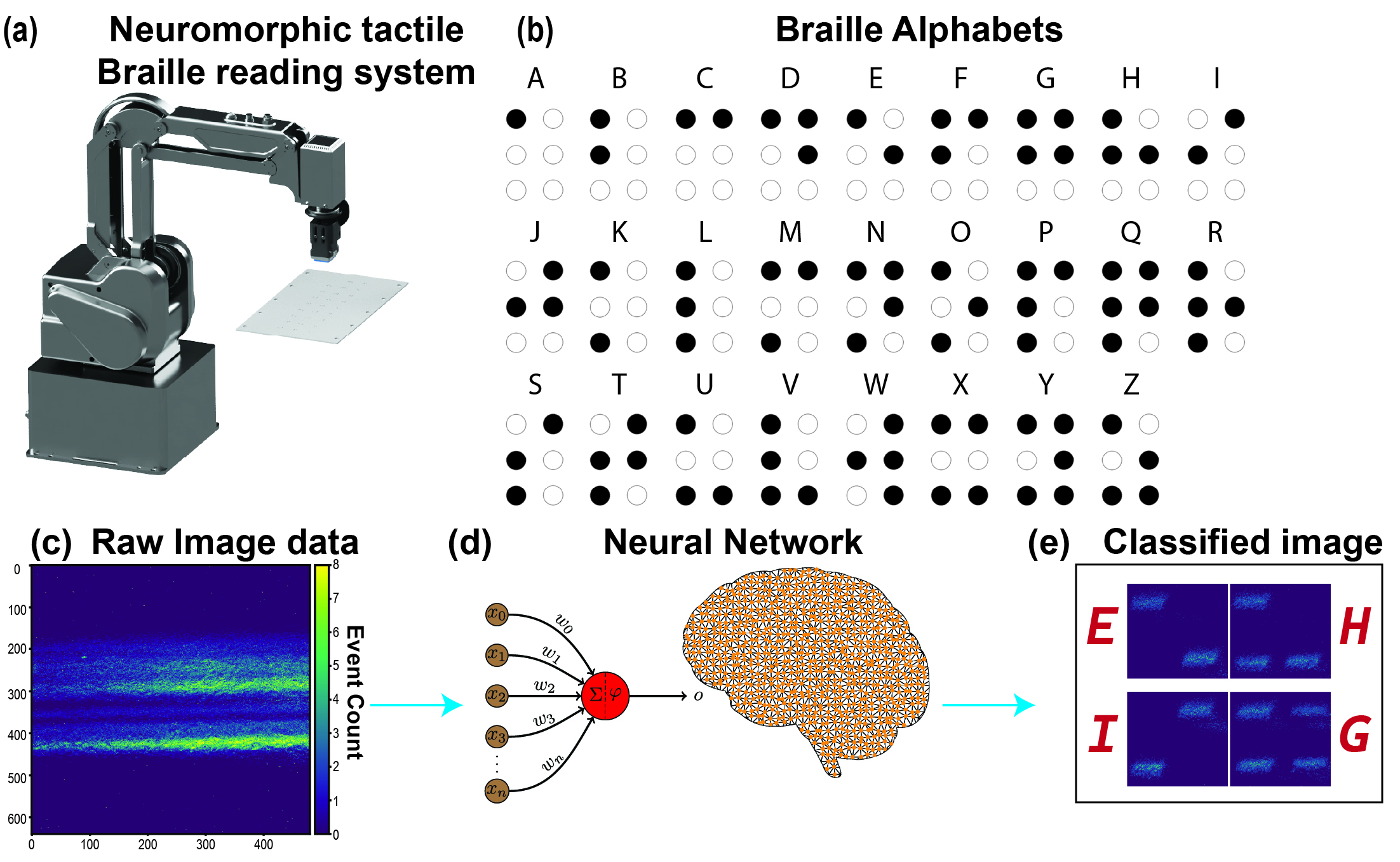}
\caption{Neuromorphic tactile-based Braille reading system: (a) A robotic arm mounted with an open-source event-based optical tactile sensor (Evetac) scanning 3D-printed Braille boards; (b) Braille alphabet layout, where black-filled circles indicate raised dots; (c) visualized spatiotemporal event streams captured by the sensor (input to the network); (d) schematic of the neural network architecture for processing the tactile input; (e) output classification results showing representative images after inference.}
\label{fig:Camera_views}
\end{figure}
Neuromorphic engineering offers a promising pathway toward addressing these challenges. Inspired by the event-based, spike-based coding of biological sensory systems, neuromorphic tactile sensors transduce mechanical stimuli into sparse, asynchronous signals that enable real-time, low-power processing~\cite{indiveri2015memory}. This paradigm is particularly well-suited for applications in prosthetics, neurorobotics, and assistive technologies, where responsiveness, efficiency, and adaptability are critical. Recently, neuromorphic vision-based tactile sensors (NVBTS) have emerged as a compelling alternative to conventional frame- \& vision-based optical tactile sensors (VBTS) ~\cite{yuan2017gelsight, ward2016tactile}, offering wider dynamic range, reduced latency, and greater energy efficiency ~\cite{gallego2020event}. Their event-driven and power-efficient design supports sparse coding and enables robust generalization under variability, making them particularly attractive for edge applications. 

In this letter, we present an end-to-end neuromorphic Braille reading system built around the open-source Evetac NVBTS, which encodes tactile events as spatiotemporal spike streams, mimicking the response of mechanoreceptors in human glabrous skin~\cite{oddo2016intraneural,funk2024evetac}. Our approach integrates Evetac with a binary segmentation network and a ResNet-based convolutional neural network trained to recognize Grade 1 Braille characters from dynamic touch sequences. The system simulates a fingertip-like scanning motion across Braille cells, capturing event streams in real time and classifying them using biologically inspired neural processing. To assess robustness under realistic conditions, we evaluated performance across varying indentation depths (Z-axis) and scanning speeds on multiple 3D-printed boards, extending validation from single characters to complete words.

Our main contributions are as follows:
\begin{itemize}
\item Generalizable Feature Learning: We propose a training framework incorporating a custom normalization and augmentation strategy ('NormAug'). This approach achieves up to $99.5\%$ character classification accuracy and ensures strong generalization across untrained contact depths ($0.2\text{--}1.5\,\text{mm}$) and varying scanning velocities.
\item Continuous Stream Decoding: We introduce a hierarchical pipeline that integrates a binary segmentation network with a high-performance character classifier, leveraging spatiotemporal features and post-processing logic. This architecture enables precise character isolation from continuous signals and robust semantic recovery, effectively compensating for tactile resolution degradation during high-speed motion.
\item High-Speed Neuromorphic System: We demonstrate an end-to-end Braille reading system based on the Evetac NVBTS capable of reading continuous words in real-time. The system maintains high fidelity ($>90\%$ word accuracy) at speeds up to $32\,\text{mm/s}$, validating the efficacy of event-driven sensing for dynamic assistive tasks.
\end{itemize}

These results demonstrate that combining neuromorphic tactile sensing with deep learning enables reliable, real-time Braille reading. More broadly, they highlight the potential of event-driven tactile perception to support assistive technologies for the visually impaired and to enable robust touch sensing in real-world conditions

\section{RELATED WORKS}

\subsection{Human Braille Perception as a Model}
Human Braille reading using tactile mechanoreceptors remains the ultimate design benchmark for artificial Braille readers. Blind trained Braille readers rely on active finger scanning and temporal–spatial integration to interpret Braille dot patterns. The fingertip-pad contact stability and the scanning dynamics are strongly predictive of Braille reading proficiency~\cite{nonaka2021structure}. The mechanoreceptors embedded in the human skin including SA-I/II (Merkel, Ruffini) and FA-I/II (Meissner, Pacinian) provide complementary encoding of static spatial structure and dynamic motion or vibration~\cite{johansson2009coding, loomis1981braille}. Scanning over the raised-dot textures sharpens spatial neural responses and improves discrimination compared with static touch~\cite{craig1999somesthesis}. These biological principles inspire artificial tactile systems that combine compliant, skin-like sensors with continuous, sliding acquisition and temporally aware processing (e.g., SNNs, neuromorphic encoders, or lightweight CNNs), reflecting recent advances in replicating SA/RA-like afferent function~\cite{pestell2022artificial}. However, despite all these advances human-like tactile robustness still remains elusive, emphasizing the urgent need for adaptive, fault-tolerant smart tactile sensing pipelines.

\subsection{Artificial Braille Reading Systems}

Artificial Braille-reading systems have progressed rapidly, from early rigid sensor grids to advanced tactile sensing technologies that support dynamic, continuous reading. In the earlier studies static pressure-sensor arrays could detect Braille engraved dot patterns, but they lacked the adaptability needed for real-world sliding or reading motion. The challenge lies in human-like tactile exploration. For example, Parth \textit{et al.} reported a high-speed robotic Braille reader using a biomimetic sliding fingertip and vision-based tactile sensing. Their system achieved up to 315 words per minute with $\approx 87\text{--}90\%$ accuracy, far exceeding typical human reading speeds~\cite{potdar2024high}. Besides pressure or static sensors, flexible optical tactile sensing has also contributed efficiently to the Braille sensing. Wang \textit{et al.} introduced a “skin-like” tactile sensor built around a flexible optical-fiber ring resonator embedded in soft PDMS. This allows the sensor to deform and respond to the raised dots as a fingertip would. Their device, when paired with neural-network based processing, achieved $\approx 98.6\%$ braille recognition accuracy, including for dynamic sliding contact~\cite{wang2025optical}. Such optical-tactile solutions are comparitively compelling because they combine the sensitivity necessary for detecting subtle dot geometry with flexibility that mimics human skin, and uses machine learning to robustly handle variations in pressure or motion~\cite{liu2023braille}. Magnetic tactile sensors, which detect Braille via magnetic field variations, represent another suitable approach~\cite{zhang2024magnetic}. These innovations enable real-time processing of dynamic Braille inputs, mimicking human touch more closely than previous designs.
Several recent reviews and experiments also confirm that combining visual and tactile modalities, or using flexible capacitive pressure sensor arrays, can yield strong Braille-recognition performance~\cite{park2023visual}. 

\subsection{Neuromorphic Tactile Sensing for Braille}
Neuromorphic tactile sensors are used because they mirror biological spike encoding and offer low-latency and energy-efficient processing which suits well for continuous Braille scanning task by mimicking the sparse, spike-based signaling of the human somatosensory system. Müller-Cleve \textit{et al.} have presented a benchmark for spatio-temporal pattern recognition on neuromorphic hardware, which recorded a dataset of Braille letters using capacitive sensors and demonstrated that event-based encoding combined with spiking neural networks (SNNs) can perform Braille classification on edge hardware with much lower power usage than conventional deep learning techniques~\cite{muller2022braille}. This hybrib bio-inspired approach is currently in the evolutionary process particularly in tactile applications as they need high resolution~\cite{lee2017discrimination}. The Braille recognition by E-skin system based on binary memristive neural network achieved up to $\approx91.25\%$ recognition accuracy using a flexible PDMS-based “e-skin” pressure sensor plus a memristor-based binary neural network demonstrating the potential of wearable, low-cost, energy-efficient Braille readers~\cite{liu2023braille}. Recently, the High‑Speed vision‑based tactile roller sensor for large surface measurements proposed a novel event-based tactile roller that scans large surfaces at high speed. The authors report braille reading performance 2.6 times faster than prior continuous tactile approaches showing how neuromorphic vision can scale to practical, real-world reading scenarios~\cite{khairi2025high}. However, still significant gaps remain before neuromorphic based tactile sensors for Braille reading become widely practical. For example: most event-based Braille classification studies use emulated event data (i.e., converting from frame-based tactile captures), because fully event-based tactile sensors are still rare~\cite{muller2022braille}. Also, existing hardware implementations, flexible e-skins or roller sensors often trade off between spatial resolution, durability (wear/tear due to sliding), and power consumption. As a result, achieving robust, high-accuracy recognition under varied real-world conditions (e.g., different Braille boards with varying patterns, varying sliding speed/pressure, environmental noise) remains still a challenge.

\section{MATERIALS AND METHODS}

\begin{figure*}[htbp]
    \includegraphics[width=\linewidth]{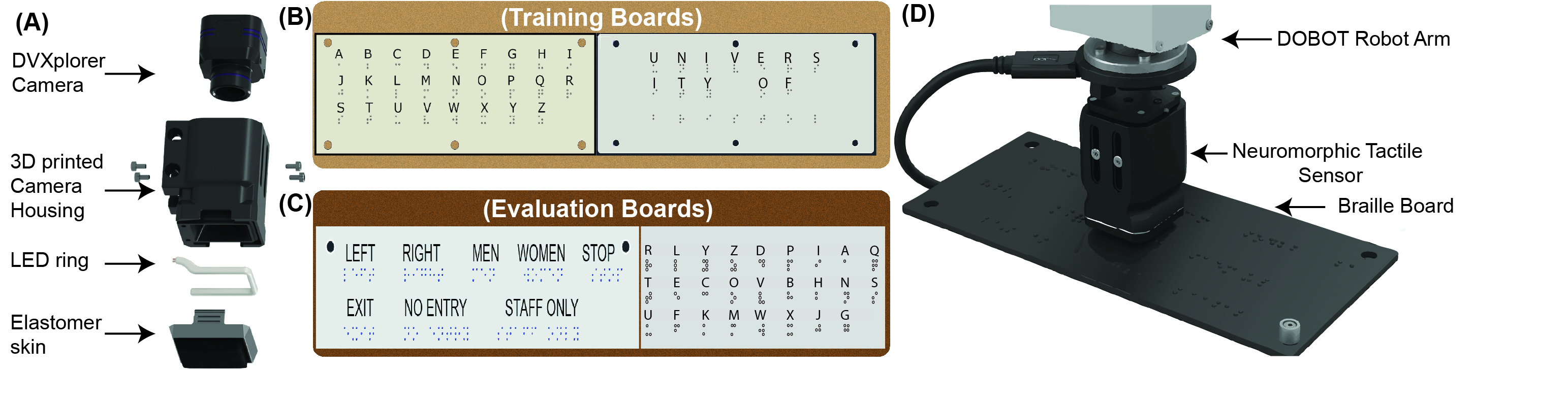}
    \caption{Experimental setup for neuromorphic Braille data collection. (A) Exploded view of the customized Evetac optical tactile sensor. From top to bottom: DVXplorer Mini event-based camera, 3D-printed camera housing, internal LED strip for illumination, and a flat GelSight Mini elastomer. (B) Training Boards: 3D-printed Braille boards used for model training. Data were collected from two boards, one containing all 26 letters of the English alphabet and two containing all characters from the phrase ``University of .......", arranged at equal spacing. (C) Evaluation Boards: 3D-printed Braille boards used to evaluate model performance. The evaluation set includes two boards containing ten commonly used words and all 26 letters of the English alphabet in random order arranged at equal distance. (D) Evetac sensor mounted on the end-effector of a DOBOT robotic arm, performing a scanning motion across the Braille boards for tactile data collection.}
\end{figure*}

\subsection{Experimental Setup}

The experimental setup consisted of a 4-degree-of-freedom (DoF) robotic arm (Dobot MG400), equipped with a customized version of the Evetac neuromorphic vision-based tactile sensor~\cite{funk2024evetac}, mounted at its end-effector. Evetac is an open-source, neuromorphic tactile sensing platform that uses an event-based camera (Inivation DVXplorer) to capture high-temporal-resolution deformations of a flexible Gelsight-like skin. Unlike traditional frame-based cameras, the event-based sensor outputs asynchronous pixel-level brightness changes, allowing for precise detection of dynamic contact events.

To adapt the Evetac sensor for fine-grained Braille recognition, we replaced the standard marker-embedded skin with a smooth, marker-free elastomer. This modification directs the sensor to capture events generated solely by the skin's deformation against Braille dots, thereby filtering out marker-related data and focusing the event stream on the immediate contact dynamics.

\subsection{Braille Boards}
For training and evaluation, four custom Braille boards were 3D printed using a Stratasys Objet Connex260 printer. Each Braille letter is made up of up to 6 hemispherical dots, with a diameter of 0.5$mm$ and 2.5$mm$ spacing between dots
(conforming to the standard dimensions set out in ANSI/NISO Z39.86-2002).

For the purpose of training, two distinct Braille boards were designed and fabricated. Each board featured different spatial configurations and arrangements of Braille characters to assess the model’s ability to identify and interpret tactile patterns under varying structural conditions.

\begin{enumerate}
    \item \textbf{Training Board 1 – Standard Alphabet Board (SAB):} 
    This board consisted of the complete 26-character English alphabet, arranged systematically across three equidistant rows (Fig. 2B). Specifically, Row~1 contained characters \textbf{A–I}, Row~2 \textbf{J–R}, and Row~3 \textbf{S–Z}. The inter-character spacing and row separation were uniform.
    \item \textbf{Training Board 2 – University Name Board (UN):} 
    This board, referred to as the \textit{UN Board}, featured an arbitrary phrase \textbf{``UNIVERSITY OF ......''}. The phrase was distributed across three rows: Row~1 – ``UNIVERSITY,'' Row~2 – ``OF,'' and Row~3 – ``.......'' The inter-character and inter-row spacing were uniform.
\end{enumerate}

For testing purposes, two additional Braille boards were fabricated to evaluate the generalizability of the model under varied spatial and linguistic conditions:

\begin{enumerate}
    \item \textbf{Test Board 1 – Randomized Alphabet Board (RAB):} 
    This board contained the same 26 Braille characters as Board~1; however, the characters were randomly distributed across the three rows instead of following alphabetical order.    
    \item \textbf{Test Board 2 (EV Board):} This board featured a curated set of ten frequently used words selected to represent symbols commonly encountered in everyday contexts (Fig. 2C). The arrangement tested the model’s performance in recognizing familiar language patterns and its adaptability to practical word-level inputs.
\end{enumerate}

\subsection{Data collection}
For data collection, the Evetac sensor performed scans at different indentation depths spanning from 0.2\,mm to 1.5\,mm.
At the start of each row, the sensor was lowered until the desired depth was achieved. Once the sensor is in contact with the board, the robotic arm moves laterally across each row over a distance of 195\,mm at one of four predefined speeds: 8\,mm/s, 16\,mm/s, 24\,mm/s and 32\,mm/s. For training, data collection was performed exclusively at 8\,mm/s \& 1.5\,mm depth to ensure consistent input conditions.
This design forces the model to generalize to the unseen depths and higher speeds used during testing.

Each indentation depth was repeated 10 times, resulting in a total of 300 trials per character (3 depths × 10 repetitions × 10 trials per row). This setup ensured that every Braille letter on the training board was scanned 300 times under controlled and repeatable conditions. Similarly, 30 trials per row were collected for each of the two test boards to ensure consistent evaluation across different layouts and word compositions.

\subsection{Data preprocessing}

We define the event stream at time \( t \) as $E(t) = \{x_i, y_i, p_i\}$,
where \( (x_i, y_i) \) represents the spatial location of the \( i \)-th event and \( p_i \in \{0, 1\} \) denotes the event polarity (ON/OFF). 
To transform raw asynchronous event data into a format suitable for convolutional neural network (CNN) processing, the events were discretized into a three-dimensional spatiotemporal heatmap representation \( H \in \mathbb{R}^{T \times H \times W} \), where \( T \) denotes the number of temporal bins, and \( H \) and \( W \) represent the spatial resolution of the sensor.
Each heatmap slice \( H_t(x, y) \) encodes the number of events that occurred at spatial location \( (x, y) \) during the \( t \)-th temporal interval, defined as:

\begin{equation}
H_t(x, y) = \sum_i \delta(t_i \in \text{bin}_t) \cdot \delta(x_i, x) \cdot \delta(y_i, y)
\label{eq:heatmap}
\tag{4}
\end{equation}

where \( (x_i, y_i, t_i) \) are the spatial and temporal coordinates of the \( i \)-th event, and \( \delta(\cdot) \) denotes the Kronecker delta function. The temporal axis was uniformly discretized using a fixed bin width of \( \Delta t = 10\,\text{ms} \).
In addition to the temporal downsampling, we also apply a spatial sub-sampling of a factor of 4 which reduces the resolution of the original images (640x480) to a resolution of 160x120.

\subsection{Image segmentation and labeling pipeline}

We next present the methods to segment the obtained image event sequences such that we can associate them to the individual characters.
This is required to create a labelled dataset that we can later train our models on.
In the first step of this procedure, we analyze the spike density over a \textbf{start window} that has a width of 10 pixels and is therefore only sensitive to the columns of individual characters.
The event counts are summed across spatial and channel dimensions to form a 1D temporal activity profile:

\begin{equation}
S(t) = \sum_{x, y, c} H_t(x, y, c)
\label{eq:sum_profile}
\tag{5}
\end{equation}

To suppress high-frequency noise while preserving temporal transitions, we smooth the raw sequence using a 1D moving average filter with a kernel size of 7, yielding $\tilde{S}(t)$.
Local maxima within $\tilde{S}(t)$ correspond to the onsets of individual Braille cell columns.

To reconstruct complete characters which may consist of either one or two columns we apply a temporal aggregation step.
Peaks occurring within a specific time window are merged into a single character instance. To account for variations in scanning speed and sensor response, we implement an adaptive search strategy for this parameter.
We iteratively expand the merging window from 300\,ms to 450\,ms until the number of detected character segments matches the known expected number of characters \( N \). This column-wise detection approach ensures the system can generalize to standard Braille spacing specifications regardless of character complexity.

Once the onset time \( t_{\text{max}}^{(i)} \) is identified for the $i$-th character, we extract sensor measurements within the temporal window \(t_{\text{max}}^{(i)}, t_{\text{max}}^{(i)} + 350ms \)]. This duration is calibrated to the slowest experimental scanning speed, ensuring that the full character trajectory remains visible within the active spatial region (pixels $20\text{--}100$). To isolate the target character from adjacent signals, we apply a spatial mask that zeros out all pixel values outside this region.
To standardize alignment, we introduce a temporal offset for single-column characters. Specifically, we shift the window start time to $t_{\text{max}}^{(i)} + 280\,\text{ms}$. This adjustment effectively centers the single column within the observation window — mimicking the spacing of a dual-column character — and ensures that subsequent characters do not intrude into the measurement frame (Fig.~4). Finally, the isolated segments are transformed into time-binned tensors and assigned class labels.
This data is subsequently used for supervised learning.

\begin{figure*}[htbp]
    \includegraphics[width=\linewidth]{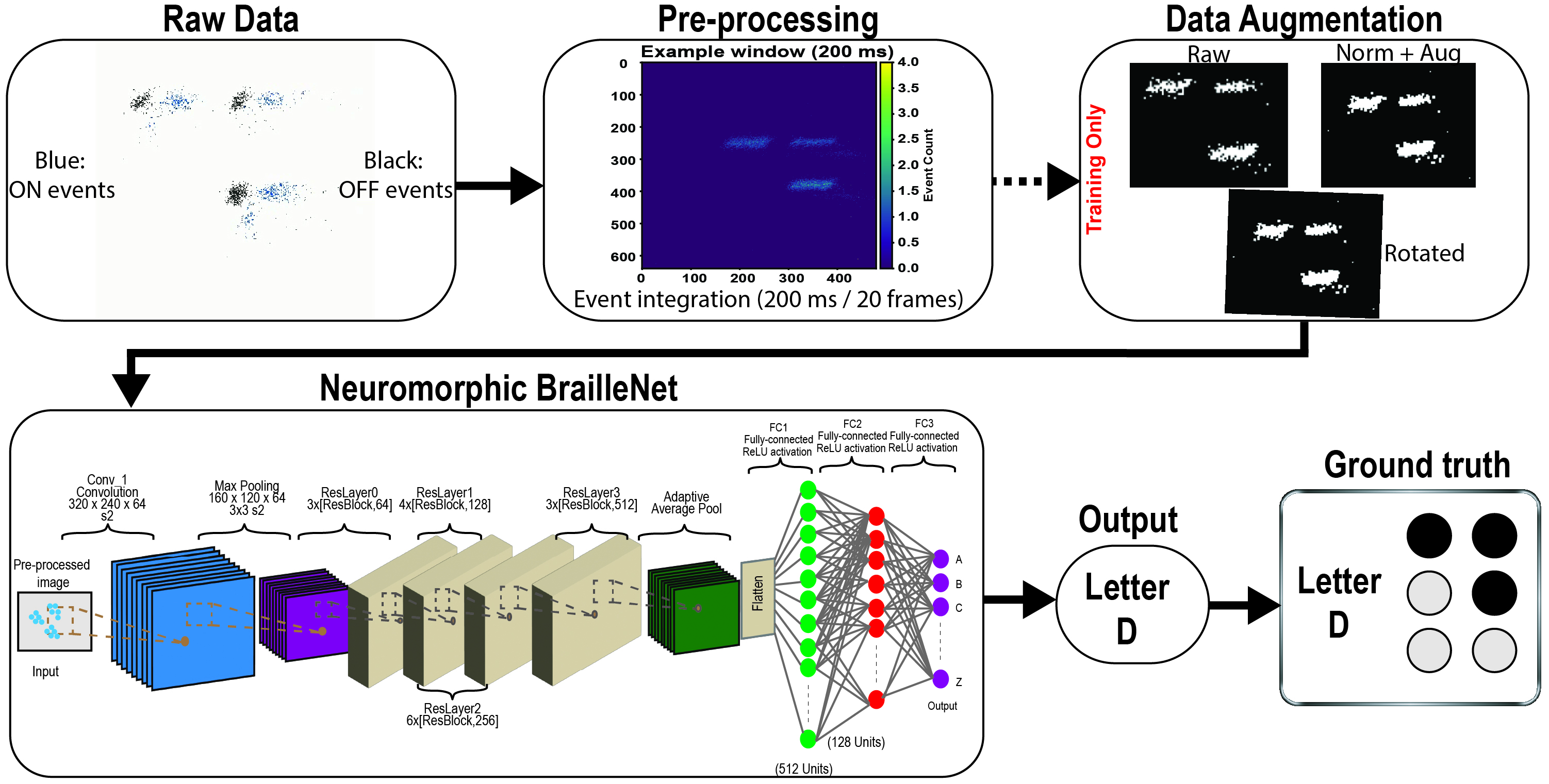}
    \caption{Schematic overview of the proposed neuromorphic tactile processing and classification pipeline based on a deep ResNet-34 architecture for Braille character recognition. Raw asynchronous tactile events are separated into ON and OFF polarities and pre-processed via spatiotemporal event integration over a 200 ms window (20 event frames) to form normalized event representations. During training, online data augmentation is applied to improve robustness to variations in contact conditions and scanning dynamics. The resulting inputs are processed by Neuromorphic BrailleNet, comprising an initial convolution and max-pooling stage followed by four residual layers (ResLayer0–ResLayer3), adaptive average pooling, and three fully connected layers. The network predicts one of 26 Braille alphabet classes, which is compared with the corresponding ground-truth Braille pattern.}
    \label{fig:network}
\end{figure*}

\subsection{Network Architecture}The proposed Braille recognition architecture (see Fig. 3) is based on a deep residual network optimized for neuromorphic event-based images. Starting with the preprocessed 1D stream of events with spatial resolution of 160x120 pixels as input, the model first applies a 7×7 convolutional layer with batch normalization and ReLU activation, followed by max pooling to reduce spatial resolution and extract low-level features (Fig.~\ref{fig:network}). Four sequential residual stages (ResLayer0–ResLayer3) then learn increasingly abstract representations, each composed of multiple residual blocks with growing channel sizes (from 64 up to 512) and progressively smaller spatial dimensions. The high-level feature maps are aggregated by an adaptive average pooling layer, producing a compact feature vector. This is fed through three fully connected layers with ReLU activation and dropout for regularization, ultimately yielding classification logits for the 26 Braille alphabet classes. This design combines efficient spatial downsampling, deep residual feature learning, and robust classification, making it well-suited for challenging, noisy Braille input data.

\subsection{Data Augmentation Strategies}
\label{sec:augmentation}

To improve robustness to variations in contact conditions and scanning dynamics, we applied data augmentation during training. Each input sample consists of a 200\,ms tactile sequence (20 event frames), which is first vertically centered by aligning its center of mass with the sensor’s midline.

With a trigger probability of 0.5, we then apply uniform spatial transformations—including rotations ($[-10^\circ, +10^\circ]$), translations (up to 10\%), and scaling (0.8 - 1.2) to model sensor noise, misalignment, and variability in sliding contact. Furthermore, to improve resilience against sensor artifacts, we introduce salt-and-pepper noise with a separate probability of 0.5, toggling event activations at a rate of $10^{-5}$.

\subsection{Learning to Read Words / Segmentation}

To transition from single-character recognition to real-time word reading, we developed a hierarchical pipeline capable of handling unsegmented data streams. For this purpose, we trained a secondary Segmentation Network — a binary classifier — tasked with distinguishing between 'character present' and 'background' states. This allows the system to query the character classifier only when valid tactile features are detected.

The segmentation model is trained on a binary dataset derived from the method in Section D. We define negative samples (non-characters) using the time windows immediately surrounding a true character, i.e., [\(t_{\text{max}}^{(i)}-170ms, t_{\text{max}}^{(i)} + 10ms \) ] and [\(t_{\text{max}}^{(i)} + 350ms -10ms, t_{\text{max}}^{(i)} + 350ms + 170ms \) ].

Finally, given the trained segmentation and character classification model, at inference time we implement a temporal filtering module to aggregate these raw detections into coherent words. To mitigate noise, we enforce a temporal consistency check: a character is only registered if detected across at least 4 consecutive frames, where 'consecutive' is defined as a gap of $<70\,\text{ms}$. Conversely, a gap exceeding $1\,\text{s}$ indicates a word boundary. These parameters act as a form of temporal averaging.

\section{RESULTS}

In this section, we evaluate the performance of our proposed approach. We begin by analyzing the classification accuracy for single Braille characters. Subsequently, we assess the model's generalization capabilities across varying indentation depths and scanning speeds. Finally, we demonstrate the system's effectiveness in a real-world setting by evaluating its performance on reading complete words.

\subsection{Braille character classification performance}

\textbf{Base performance - single Braille characters at constant speed and depth.}
As an initial evaluation of our BrailleNet method, we train the system by gathering data on the SAB and UN boards at a constant speed of 8\,mm/s and indentation depth of 1.5\,mm as described in the methodology. The models are tested on a different board containing all alphabet characters in a random order (RAB).

Table~\ref{tab:base_performance_braille_diff_configurations} details the performance of various model configurations under different data augmentation strategies. Even when evaluated at the same depth and sliding speed as the training set, our pre-processing pipeline proves highly effective, boosting accuracy from a baseline of $\approx 86\%$ on raw data to $99.54\%$ with full normalization and augmentation (Sec.~\ref{sec:augmentation}). This substantial gain highlights the inherent variability in the data—likely attributable to minor manufacturing tolerances in the 3D-printed boards and fluctuations in sensor-character alignment—and confirms that augmentation successfully compensates for the limited training data.

\begin{table}[t]
\centering
\caption{Braille character classification performance under different configurations. For the performance of the NormAug model see also Fig.~\ref{fig:confusion_matrix}.}
\begin{tabular}{lccc}
\toprule
\textbf{Configuration} & \textbf{Accuracy (\%)} & \textbf{F1-score} & \textbf{Loss} \\
\midrule
Raw        & 85.62 & 0.8218 & 2.0282 \\
Norm       & 94.90 & 0.9441 & 0.2536 \\
Aug        & 93.22 & 0.9178 & 0.1751 \\
NormAug    & 99.54 & 0.9954 & 0.0340 \\
\bottomrule
\end{tabular}
\label{tab:base_performance_braille_diff_configurations}
\end{table}

\begin{figure}[htbp]
    \centering
    \includegraphics[width=.8\columnwidth]{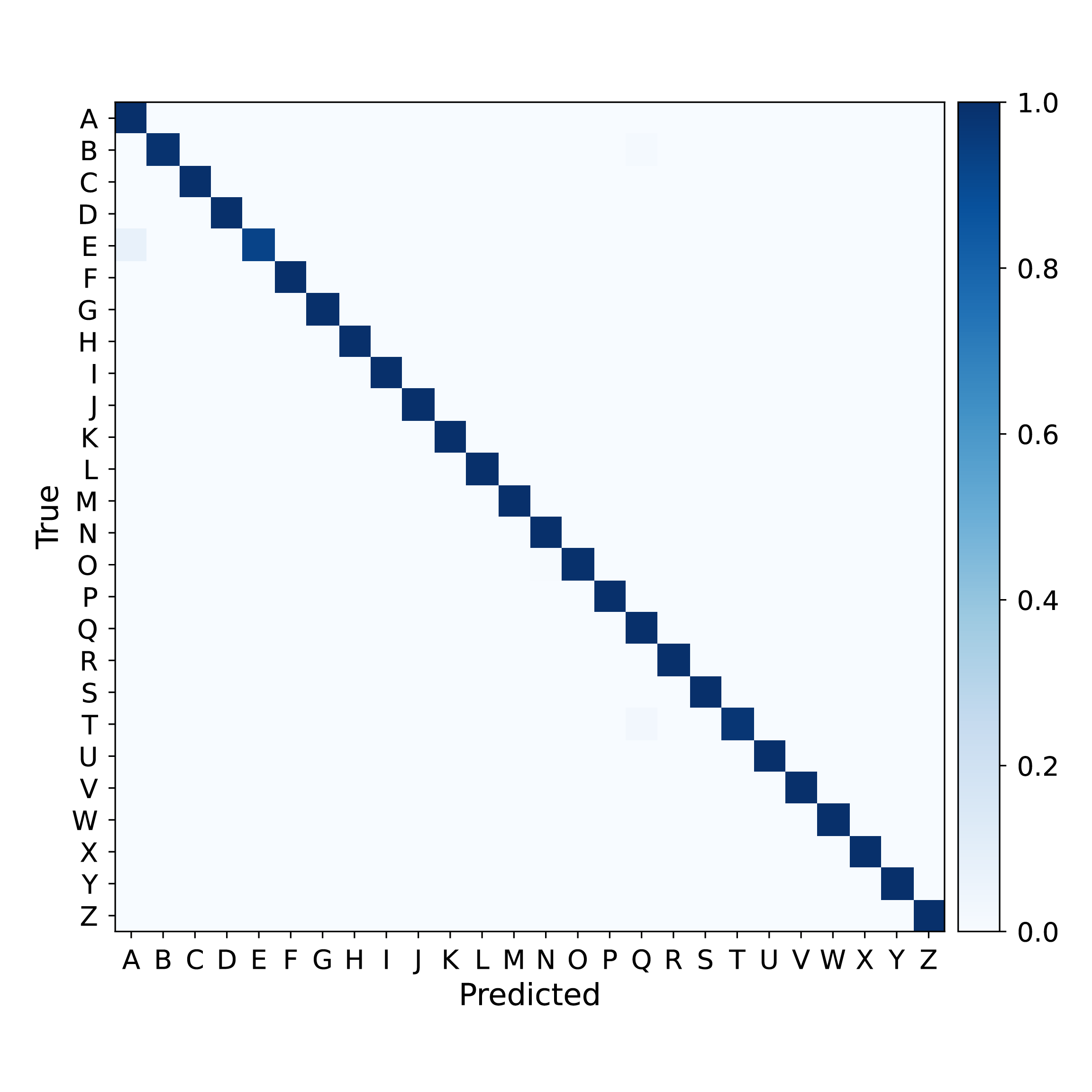}
    \caption{Classification accuracy of the proposed ResNet-based Braille recognition model for random alphabet board. (99.54\% accuracy)}
    \label{fig:confusion_matrix}
\end{figure}

Furthermore, the confusion matrix for the NormAug model (Fig.~\ref{fig:confusion_matrix}) illustrates robust classification performance. The only minor notable error occurs between the true label 'E' and the predicted label 'A'. This confusion, though rare, likely stems from the sensor occasionally failing to register the middle-right dot of the 'E' character.

Overall, these results validate the model's accuracy under controlled conditions, establishing a strong baseline for the subsequent evaluation in more dynamic and challenging scenarios.

\textbf{Generalization Across Varying Contact Depths.}
We evaluate the robustness of BrailleNet by testing classification performance across a spectrum of indentation depths (0.2-1.5\,mm). Importantly, the results reported here utilize the same model from the previous section—trained exclusively at 1.5\,mm depth — to rigorously test generalization to unseen contact conditions.
To quantify the variation in signal characteristics, we analyzed the average event density per Braille dot. As indentation depth increases, the sensor response intensifies significantly: relative to the baseline at 0.2\,mm, the event count increases by 45\%, 71\%, and 91\% at depths of 0.6\,mm, 1.0\,mm, and 1.5\,mm, respectively.

\begin{figure}[htbp]
    \centering
    \includegraphics[width=\columnwidth]{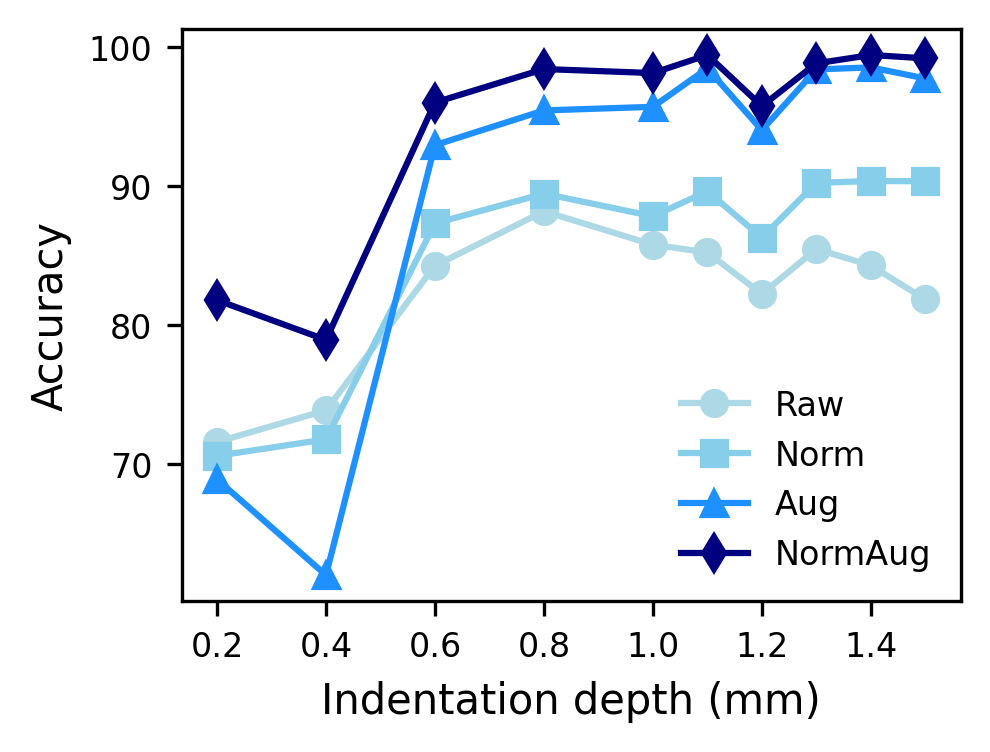}
    \caption{Classification accuracy of the proposed ResNet-based Braille recognition model across indentation depths from 0.2 to 1.5\,mm.}
    \label{fig:depth}
\end{figure}

Performance generally improves as the test depth approaches the training depth of 1.5\,mm, with a notable rise in performance between 0.4\,mm and 0.6\,mm (Fig.~\ref{fig:depth}) . Crucially, the NormAug strategy consistently outperforms other configurations across all depths. Our analysis reveals a distinct role for each component: while data augmentation alone provides a reliable boost at depths similar to the training set, the normalization step becomes critical for handling larger depth discrepancies. Consequently, only the combination of normalization and augmentation (NormAug) ensures robust performance across the full range of physical variations.

\subsection{Model Performance on a Braille Reading Board with Daily-Living Vocabulary}

Building on the robust performance of the NormAug model, we extend our evaluation from isolated characters to continuous word reading. We utilize a dedicated evaluation board containing two rows of text (comprising four and five words, respectively) and deploy the full system, integrating both the character classification and segmentation networks.

We conducted 30 scanning trials per row across varying indentation depths. To analyze performance at different levels of granularity, we define four hierarchical metrics. Importantly, these metrics are conditional: each subsequent metric is evaluated only on the subset of data that satisfied the preceding segmentation criteria.
The evaluation criteria are:
(1) number of correct words per line, i.e., on how many of the slides across the letters the method detected the overall correct number of words per line, (2) number of correct letters per word, for how many of all the words, we predicted and detected the correct number of letters per word, (3) the number of correct words, (4) number of correct letters if for the respective word we predicted the right number of letters.

We evaluate the models using the NormAug augmentation. Additionally, we assess the impact of adding a spell checker (SC) to correct minor classification errors. We selected the \texttt{textblob} Python library for this task specifically because it operates statelessly; this guarantees that the correction of one trial is not influenced by the history of previous scans over the same text rows.

\begin{figure}[htbp]
    \centering
    \includegraphics[width=0.9\columnwidth]{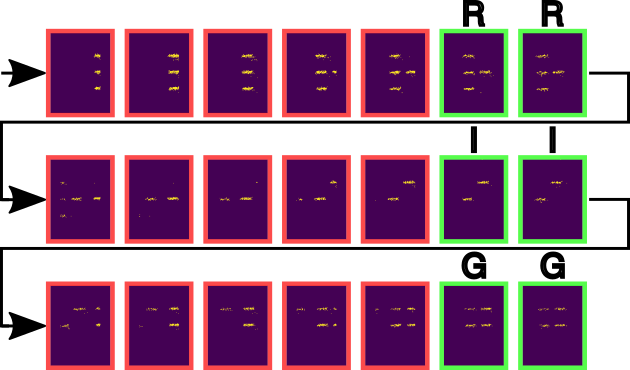}
    \caption{Illustration of the segmentation challenges in continuous Braille reading. The figure displays the temporal evolution of sensor measurements during a scan of the word "RIGHT" (first row of the evaluation board). Red-bordered frames represent transition states filtered out by the segmentation model, preventing classification when no distinct character is fully visible. Green-bordered frames indicate valid instances where the character classifier is triggered. As demonstrated, the segmentation model is critical for rejecting ambiguous intermediate states; without it, the classifier would evaluate transition frames between characters, leading to spurious detections.}
    \label{fig:importance_segmentation}
\end{figure}

Figure~\ref{fig:importance_segmentation} illustrates the critical role of the segmentation model in filtering the raw camera stream. By isolating valid character windows from noisy transition frames, the segmentation module ensures that the classification network is evaluated only on clean data. Without this pre-filtering, spurious predictions on intermediate frames would render accurate word identification impossible.

\begin{table}[!t]
    \renewcommand{\arraystretch}{1.25}
    \caption{Performance of the model at different sliding speeds}
    \centering
    \label{table_preliminary_res_words_diff_speeds}
    \begin{tabularx}{\columnwidth}{lXXXX}
        \toprule
\thead{Speed} &
\thead{Words \\ per Line} &
\thead{Letters \\ per Word} &
\thead{Correct \\ Words } &
\thead{Correct \\ Letters} \\
        \midrule\midrule
        \thead{8\,mm/s} & 0.916 & 0.883 & 0.876 & 0.882 \\
        \thead{8\,mm/s \\ w spell} & 0.916 & 0.903 & 0.903 & 0.903 \\
        \thead{16\,mm/s} & 1.0 & 0.9 & 0.866 & 0.889 \\
        \thead{16\,mm/s \\ w spell} & 1.0 & 0.983 & 0.97 & 0.976 \\
        \thead{24\,mm/s} & 0.933 & 0.773 & 0.753 & 0.768 \\
        \thead{24\,mm/s \\ w spell} & 0.933 & 0.913 & 0.913 & 0.913 \\
        \thead{32\,mm/s} & 1.0 & 0.728 & 0.685 & 0.713 \\
        \thead{32\,mm/s \\ w spell} & 1.000 & 0.986 & 0.971 & 0.978 \\
        \bottomrule
    \end{tabularx}
\end{table}

Moreover, Table~\ref{table_preliminary_res_words_diff_speeds} details the results across various scanning speeds. With the spell checker enabled, accuracy remains robust ($>90\%$) across all conditions, showing only minor fluctuations attributable to signal noise or motion variability.
However, the results without the spell checker reveal a performance bottleneck at higher velocities. Specifically, at speeds $\geq 24\,\text{mm/s}$, we observe a degradation in the segmentation network's ability to correctly delineate individual characters. This suggests that while the current system is robust, the segmentation network would benefit from a more diverse training set to better handle the signal dynamics of high-speed scanning. Nevertheless, the complete end-to-end system demonstrates stable and reliable word-level Braille reading.

\section{DISCUSSION}

The findings of this work demonstrate the potential of neuromorphic tactile sensing for continuous Braille recognition, offering significant advantages over conventional frame-based and discrete scanning approaches. By leveraging event-based tactile sensing, our system achieved near-perfect recognition accuracy at standard indentation depths and maintained strong performance across a range of scanning speeds. This performance aligns with prior reports on event-driven tactile sensing for spatiotemporal pattern recognition~\cite{muller2022braille, zhuang2022neuromorphic}, while extending these works to continuous, naturalistic reading conditions that more closely mimic human Braille reading. The results highlight the suitability of neuromorphic perception for real-time assistive applications, where latency and computational efficiency are critical~\cite{indiveri2015memory}.

A key contribution of this work is a systematic evaluation of indentation depth, a parameter that strongly affects the visibility and salience of Braille dot impressions. The depth–accuracy results (Fig~\ref{fig:depth}) reveal two trends. First, performance improves substantially as indentation increases from 0.2 to 0.6 mm, attributable to the sharper, more stable tactile signatures generated at moderate depths. Second, the NormAug configuration consistently provides the highest accuracy across depths, reaching more than 95\% at all depths apart from 0.2\,mm. 
These findings show that combining spatial normalisation with targeted augmentation creates representations that are robust to the variability commonly found in real Braille materials, including worn embossing, low-relief printing, and differences in substrate stiffness.

Performance across scanning speeds demonstrates the system's robustness to temporal dynamics. As detailed in Table~\ref{table_preliminary_res_words_diff_speeds}, the full pipeline (with spell checking) maintains a word accuracy above $90\%$ across the entire tested range ($8\text{--}32\,\text{mm/s}$).The system achieves peak performance at $16\,\text{mm/s}$ and $32\,\text{mm/s}$, with accuracies reaching $\approx 97\%$. Notably, the robustness at $32\,\text{mm/s}$ is achieved despite a significant drop in raw sensor classification accuracy (Table~\ref{table_preliminary_res_words_diff_speeds}, row 7).
This aligns with findings that high-speed sliding reduces tactile resolution due to integration limits~\cite{johansson2009coding}; however, our results show that the combination of neuromorphic sensing and language post-processing effectively compensates for this loss.

Beyond technical performance, this work has broader implications for assistive technologies. The demonstrated generalization across multiple Braille boards and daily-living vocabulary suggests applicability to real-world Braille reading devices. Unlike vision-based pipelines that require high-resolution imaging and intensive preprocessing~\cite{potdar2024high}, the neuromorphic framework provides low-latency, power-efficient inference that could be deployed on portable or embedded hardware. Such systems could benefit not only visually impaired users but also robotic applications in texture recognition, material classification, and slip detection~\cite{oddo2016intraneural}, expanding the role of neuromorphic tactile sensing in broader domains of robotic manipulation and haptics.

Despite these promising results, several limitations remain. The present work was restricted to Grade 1 Braille, and scaling to Grade 2 and contracted Braille will require integration of higher-level language models for contextual disambiguation. Real-world usability also demands further validation against environmental variables, including prolonged operation, diverse surface materials, and potential interference from ambient lighting in optical-tactile setups.
Additionally, while event-driven systems are theoretically power-efficient, specific energy consumption metrics were not quantified here and will be benchmarked in future embedded implementations. Addressing these challenges potentially through hybrid neuromorphic–deep learning models or recurrent spiking architectures will be critical for translating laboratory performance into robust, user-ready assistive systems.

\section{CONCLUSION}

This work demonstrates that neuromorphic tactile sensing enables accurate, real-time Braille reading under continuous, naturalistic scanning conditions. By combining an event-based tactile sensor with segmentation and residual networks, the proposed system achieves robust word-level performance across varying indentation depths and scanning speeds. These results highlight the potential of neuromorphic touch as a scalable, low-latency foundation for assistive Braille readers and broader tactile perception tasks in robotics.

\addtolength{\textheight}{-12cm}   




\section*{ACKNOWLEDGMENT}


\bibliographystyle{IEEEtran}
\bibliography{main.bib}

\end{document}